\documentclass[sn-mathphys,Numbered]{sn-jnl}% Math and Physical Sciences Reference Style
\usepackage{graphicx}%
\usepackage{multirow}%
\usepackage{amsmath,amssymb,amsfonts}%
\usepackage{amsthm}%
\usepackage{mathrsfs}%
\usepackage[title]{appendix}%
\usepackage{xcolor}%
\usepackage{textcomp}%
\usepackage{manyfoot}%
\usepackage{booktabs}%
\usepackage{algorithm}%
\usepackage{algorithmicx}%
\usepackage{algpseudocode}%
\usepackage{listings}%
\usepackage{booktabs}
\usepackage{graphicx}
\usepackage{algorithm}
\usepackage{algpseudocode}
\usepackage{comment}
\usepackage[utf8]{inputenc}
\usepackage{amsmath}
\usepackage{tabularx}
\usepackage{caption}

\theoremstyle{thmstyleone}%
\theoremstyle{thmstyletwo}%

\theoremstyle{thmstylethree}%

\begin{document}
\fbox{This article is under review. Content starts from the next page}
\title[Article Title]{Randomize to Generalize: Domain Randomization for Runway FOD Detection}
% Author Orchid ID: enter ID or remove command
\newcommand{\orcidauthorA}{0000-0001-8763-3334} % Add \orcidA{} behind the author's name
\newcommand{\orcidauthorB}{0000-0003-2763-2094} % Add \orcidB{} behind the author's nam


\author*[1]{\fnm{Javaria} \sur{ Farooq}}\email{jfarooq.ms14avecae@student.nust.edu.pk}

\author[1]{\fnm{Nayyer } \sur{Aafaq}}

\author[2]{\fnm{Muhammad Khizer Ali } \sur{Khan}}

\author[1]{\fnm{Ammar } \sur{Saleem}}

\author[3]{\fnm{Muhammad Ibraheem } 
\sur{Siddique}}

\affil*[1]{College of Aeronautical Engineering, National University of Sciences and
Technology, Islamabad, Pakistan}

\affil[2]{Khalifa University of Science and Technology, Abu Dhabi, UAE}

\affil[3]{Mohamed Bin Zayed University of Artificial Intelligence, Abu Dhabi, UAE}


\abstract{Tiny Object Detection is challenging due to small size, low resolution, occlusion, background clutter, lighting conditions and small object-to-image ratio. Further, object detection methodologies often make underlying assumption that both training and testing data remain congruent. However, this presumption often leads to decline in performance when model is applied to out-of-domain(unseen) data. Techniques like synthetic image generation are employed to improve model performance by leveraging variations in input data. Such an approach typically presumes access to 3D-rendered datasets. In contrast, we propose a novel two-stage methodology Synthetic Randomized Image Augmentation (SRIA), carefully devised to enhance generalization capabilities of models encountering 2D datasets, particularly with lower resolution which is more practical in real-world scenarios. The first stage employs a weakly supervised technique to generate pixel-level segmentation masks. Subsequently, the second stage generates a batch-wise synthesis of artificial images, carefully designed with an array of diverse augmentations. The efficacy of proposed technique is illustrated on challenging foreign object debris (FOD) detection. We compare our results with several SOTA models including CenterNet, SSD, YOLOv3, YOLOv4, YOLOv5, and Outer Vit on a publicly available FOD-A dataset. We also construct an out-of-distribution test set encompassing $800$ annotated images featuring a corpus of ten common categories. Notably, by harnessing merely  $1.81\%$ of objects from source training data and amalgamating with 29 runway background images, we generate  $2227$ synthetic images. Subsequent model retraining via transfer learning, utilizing enriched dataset generated by domain randomization, demonstrates significant improvement in detection accuracy. We report that detection accuracy improved from an initial  $41\%$ to $92\%$ for OOD test set.}

%%================================%%
%% Sample for structured abstract %%
%%================================%%

\keywords{Object Detection; Tiny Object Detection, FOD Detection, Deep Learning, Synthetic Images}


\maketitle

\section{Introduction}\label{sec1}

In real-world scenarios, the weather, environment, and lighting conditions demonstrate substantial variations throughout the span of a year. Moreover, objects that belong to the same categories exhibit significant differences in terms of their structure, visual characteristics, and types. These factors pose challenges for object detectors as the test data or real-world data deviates significantly from the training datasets. 

Modern detectors have demonstrated remarkable performance in the context of training and testing with data from a single benchmark \cite{zhang2020semantics}. However, various real-world application scenarios require consistent and reliable performance even when presented with input data from diverse distributions, including those that deviate from the distribution of the training data \cite{liu2021survey,shen2021towards,zaidi2022survey,zou2023object}. 

Given the limitations of pre-collected data in capturing all potential scenarios in real-world applications, the capacity of models to effectively adapt to unfamiliar distributions becomes a critical factor for detectors. As a result, the training and testing of detectors only using data obtained from a particular dataset lack practical value in terms of assessment. Consequently, these detectors may exhibit significant degradation in performance even when subjected to slight disruptions \cite{chen2021robust, hasan2021generalizable}. This scenario prompts a crucial investigation: How can we improve the capacity of contemporary detectors to generalize in real-world scenarios that are unfamiliar? 

The identification of Foreign Object Debris (FOD) on runways is one of the real-world applications that experiences performance deterioration due to the diverse forms, sizes, and classifications of FOD. The combination of these factors, together with the presence of environmental variables on runways, contributes to an overall reduction in the model's performance when deployed in real-world scenarios.

Foreign Object Debris (FOD) refers to objects that are discovered in close proximity to an aircraft and possess the potential to inflict severe harm against the aircraft itself or the individuals present on board \cite{chauhan2020review}. FOD has distinct characteristics that set it apart from conventional objects, since FOD items are often diminutive in size, exhibit a wide range of appearances, and may even possess deformed forms. While commercial automated FOD detection technologies provide advantages such as increased speed and reduced human error compared to manual approaches, their high cost remains a significant barrier to widespread implementation in airports worldwide.

The conventional approaches for FOD identification mostly include millimeter-wave radars \cite{zhong2021fod}\cite{yonemoto2018two}, optical image sensors \cite{cao2016foreign,cao2017detecting,cao2018region}, and hybrid target detection technology utilizing several sensors \cite{hong2018experiment}\cite{yuan2020research}. Among the standard ways, optical image sensor technology is characterized by its relatively inexpensive cost.  The FOD detection system is composed of 8-10 optical image sensors, which are strategically positioned at a distance of 150 meters from the center line of the runway. The number of sensors is determined depending on the length of the runway. When Foreign Object Debris (FOD) is detected, an alarm is activated and a visual alert is transmitted to the Air Traffic Controller. This is done in order to facilitate the timely and appropriate disposal of the FOD.

The evaluation of deep learning computer vision algorithms for the detection and classification of Foreign Object Debris (FOD) using optical sensors is conducted on various FOD datasets. These algorithms primarily consist of two-stage object detectors, including CNN with Region Proposal Network (RPN) \cite{cao2016foreign}, CNN with improved RPN \cite{cao2018region}, and CNN with Faster RCNN \cite{liu2018fod}. Additionally, there are one-stage object detectors such as Lighter Network SSD \cite{cao2017detecting}, SSD \cite{munyer2021fod}, and YOLOv3 \cite{cao2018region}\cite{papadopoulos2021uav}. Previous studies \cite{cao2016foreign}\cite{cao2017detecting} \cite{cao2018region} \cite{liu2018fod} \cite{papadopoulos2021uav} have mostly concentrated on training and evaluating FOD detection techniques using proprietary datasets that are not publicly accessible. These datasets typically have a limited number of FOD categories, ranging from 2 to 5. The FOD in Airport (FOD-A) dataset\cite {munyer2021fod}, consisting of 31 categories of FOD, has been investigated in recent literature to tackle the challenge of FOD detection \cite{noroozi2023towards} \cite{munyer2022foreign}.

In our previous study, we conducted an empirical evaluation comparing the performance of the state-of-the-art anchor-free object detection approach, specifically CenterNet, with anchor-based object detection approaches, namely scaled YOLOv4, YOLOv5, and SSD. Our findings indicated that among the anchor based object detectors, the YOLOv5m model \cite{GC}, due to its compact model size and default augmentation techniques, achieved the most favorable balance between speed and accuracy for the FOD-A dataset. The utilization of enhanced mosaic augmentation, which is an extension of cut-mix augmentation, inside the YOLOv5 framework results in a reduction of the network's focus on the background. This reduction is achieved by combining four images together in a tiled manner throughout the training process. The YOLOv5m model, which was trained on the FOD-A dataset, demonstrated a detection accuracy exceeding 90\% when evaluated on the FOD-A test set. However, during a qualitative assessment of the model's performance on FOD images that were not included in the training data, it was observed that the model not only failed to detect FODs but also produced false alarms by erroneously associating foreground elements with the background. The ability to generalize a model to unseen data is a fundamental performance criterion evaluating the effectiveness of FOD identification systems, given the practical implications of FOD detection in real-world scenarios. 

One plausible rationale for the limited generalization observed in deep learning models for FOD identification is their implicit assumption that the known FOD categories previously identified in airport settings represent the complete range of FOD objects. In practice, airport settings possess distinct qualities such as their expansive nature, dynamic atmosphere, and the prevalence of many unpredictable elements. Consequently, the set of foreign object debris (FOD) inside these environments remains limitless due to the ongoing discovery of novel sorts of foreign objects. In addition, the identification of foreign objects is also hindered by the fluctuating meteorological circumstances, including but not limited to sunny or rainy weather, as well as the assortment of airport surface materials, such as concrete and asphalt \cite {noroozi2023towards}. 
 
Large-scale benchmark datasets, such as ImageNet \cite{deng2009imagenet, russakovsky2015imagenet} and Microsoft COCO \cite{lin2014microsoft}, possess a generic character. Consequently, their primary utility lies in facilitating pre-training (transfer learning) of models for particular industrial applications, such as autonomous FOD detection. The unique challenge faced in the industrial context is the physical acquisition of a dataset accompanied by annotations. This task becomes notably arduous in scenarios involving the identification of FOD when occurrences of such events are rare. Considering the intricate and diverse nature of FOD items, the limited availability of FOD datasets, and the challenges associated with collecting FOD data, employing artificial augmentation techniques to produce synthetic images proves to be a suitable approach for improving FOD datasets. 

\subsection{Research Focus \& Contributions}
The main focus of this study is the development of a novel framework that comprises various components designed particularly to address the unique aspects of the FOD detection problem. Based on our comprehensive assessment, we have identified the most effective amalgamation that meets the necessary criteria for doing automated FOD inspections on runways. The research study has made several distinct contributions, which may be succinctly stated as follows.
\vspace{2mm}
\begin{itemize}
	\item  
 This study involves the creation of an out-of-distribution (OOD) test set consisting of 800 authentic runway foreign object debris (FOD) images, encompassing 10 prevalent categories alongside the FOD-A dataset for the quantitative evaluation of cross-dataset generalization 
	\vspace{2mm}
	\item 
 This work introduces a unique two-step methodology known as Synthetic Randomized Image Augmentation (SRIA) to enhance the generalization capabilities of the YOLOv5m model. The proposed strategy aims to improve the dataset without requiring further annotation efforts.
	\vspace{2mm}
		\item The research proposes an algorithm to generate batch-wise synthetic FOD images by randomizing augmentations and background substitution,
	\vspace{2mm}
	\item 	
 In addition, this study aims to assess the efficacy of transfer learning and mix-up augmentation techniques in enhancing the generalization capabilities of the model.
\end{itemize} 
\vspace{2mm}
The contribution of this paper is threefold. First, we prepare a test set comprising actual runway FOD images to serve as a baseline for future research. Secondly, we propose a two-step approach in which the first step employs weak supervision to get segmentation masks saving annotation effort, and the second step proposes a domain randomization-based approach to induce variation in synthetic data generation. We also empirically evaluate the improvement in generalization of the YOLOv5 model from mix-up augmentation and transfer learning. During the evaluation phase, we compare different models SSD \cite {munyer2021fod}, YOLOv3 \cite {munyer2021fod}, CenterNet , YOLOv4 , Outer Vit \cite{munyer2022foreign}, YOLOv4-csp w/Augmentation \cite{noroozi2023towards} on a publicly available dataset of FOD (Foreign Object Debris) images \cite{munyer2021fod} as well as an unseen test set collected on runway under varying lighting conditions (OOD test set). The results of the experiment demonstrate that our proposed FOD detection approach SRIA coupled with YOLOv5 object detector outperforms all compared approaches in terms of average precision and recall rate.

The subsequent sections of the paper are organized in the following manner: Section 2 presents an overview of fundamental terminologies. Section 3 provides an introduction to the relevant literature while Section 4 outlines the proposed framework for detecting Foreign Object Debris (FOD), which includes a two-step data-driven methodology aimed at improving the performance of the FOD detection model. Section 4 further elaborates on the conducted experiments, offering comprehensive evaluation and details followed by Section 5 presenting an analysis and explanation of the experimental results. Section 6 of the study provides a reflective analysis of the acquired insights derived from the research conducted, while also highlighting potential avenues for future research.
\section{Preliminaries}
\label{sec:mat&meth}
\subsection{Domain Generalization}  
\label{sec:domain randomization}
Performance degradation of object detection models on out-of-domain (unseen) data originates from the assumption that domain of training and test data is the same \cite{seemakurthy2022domain}, which is not the case for most of real-world machine vision applications. Although computer vision dataset are meant to be representative of the real world, Torralba et al. \cite{torralba2011unbiased} established that most of the datasets do not represent all aspects of the visual world completely through a toy experiment. Domain generalization refers to learning domain-invariant features aiming to make a model generalize well on unseen target domains by suppressing domain-specific features\cite{seemakurthy2022domain}. The aim of domain generalization is to learn a generalizable predictive function: 
\begin{equation}
\begin{aligned}
h: x \rightarrow y
\end{aligned}
\end{equation}

acquiring M training domains to get minimum prediction error in $S_{\text {test }}$domain  (not accessible during training) while ensuring that:
\begin{equation}
\begin{aligned}
P_{x y}^{\text {test }} \neq P_{x y}{ }^i \text { where } \mathrm{i} \in\{1 \ldots . M\}
\end{aligned}
\end{equation}

It differs from domain adaptation as in domain adaptation, targets without labels are accessible during training and the network is tested on a known target domain \cite{yue2019domain}. Domain generalization refers to a problem where training and testing data are from the same classes but different distributions. It makes it different from zero-shot learning where the aim is to maximize classification performance on classes that are not seen during training\cite{wang2022generalizing}. 
\subsection{Data Augmentation}  
Data augmentation is a methodology utilized to enhance a dataset by implementing a range of operations, including scaling, rotation, translation, and similar transformations, to the preexisting images contained within the dataset. The main goal of data augmentation is to improve the generalization \cite{yue2019domain} capacity of machine learning models. This is accomplished by providing the model with guidance to prioritize crucial characteristics while disregarding less pertinent or superfluous information throughout the process of object recognition and classification. The utilization of this methodology enhances the efficacy and resilience of the model in effectively addressing a wide range of complex real-world scenarios.

\subsection{Mix-up Augmentation}
Mix-up augmentation augmentation improves the robustness and generalization of the model by empirical risk minimization. The main concept of mix-up is to mix features and corresponding labels to reduce over-confidence in the network regarding the relationship between features and labels. Mix-up constructs virtual examples from two random training examples through linear interpolation with very little computational overhead. Mathematically mix-up augmentation is stated as:
\begin{equation}
\begin{aligned}
& \tilde{x}=B \odot x_1+(1-B) \odot x_2 \\
& \tilde{y}=\lambda y_1+(1-\lambda) y_2
\end{aligned}
\end{equation}

where $x_{\text {1 }}$ and $x_{\text {2}}$  are the original input images,  $y_{\text {1 }}$ and  $y_{\text {2 }}$  are one-hot label encodings, B  is  mixing mask matrix, I is an identity matrix while $\odot$ depicts element-wise matrix  multiplication

\subsection{Transfer Learning}
%Transfer learning refers to learning low level features such as edges and other geometric shapes from pre-trained models of large datasets since these features are not related to a image content in target dataset. Subsequently, data from target dataset is used to fine-tune the model \cite{knoll2019assessment}.
Transfer learning is a complex machine learning technique that involves using pre-trained models on large datasets to acquire low-level characteristics, such as edges and geometric forms. The primary stage of the learning process concentrates on elements of the data that are not reliant on the particular content of the dataset being analyzed. Subsequently, the model undergoes additional refinement through the utilization of data derived from the target dataset itself \cite{knoll2019assessment}.

The fundamental objective of transfer learning is to elevate the performance of a model when applied to a distinct but related task in comparison to the original task for which the model was initially trained \cite{wang2022generalizing}. In essence, it leverages the knowledge and representations gained from the source task to expedite and enhance the learning process for the target task, thus capitalizing on the domain knowledge and features acquired during the initial training.

\subsection{Domain Randomization} Domain randomization focuses on getting generalized features relying on variation of input data using synthetically generated images. It is a complementary approach to DG \cite{seemakurthy2022domain}. The goal of domain randomization is to generate synthetic data with enough variations that the model perceives the real world as another variation which bridges the reality gap in object detection \cite{yue2019domain}

\begin{figure}[htbp]
%\vspace{-5mm}	
		\centerline{\includegraphics[width =\textwidth]{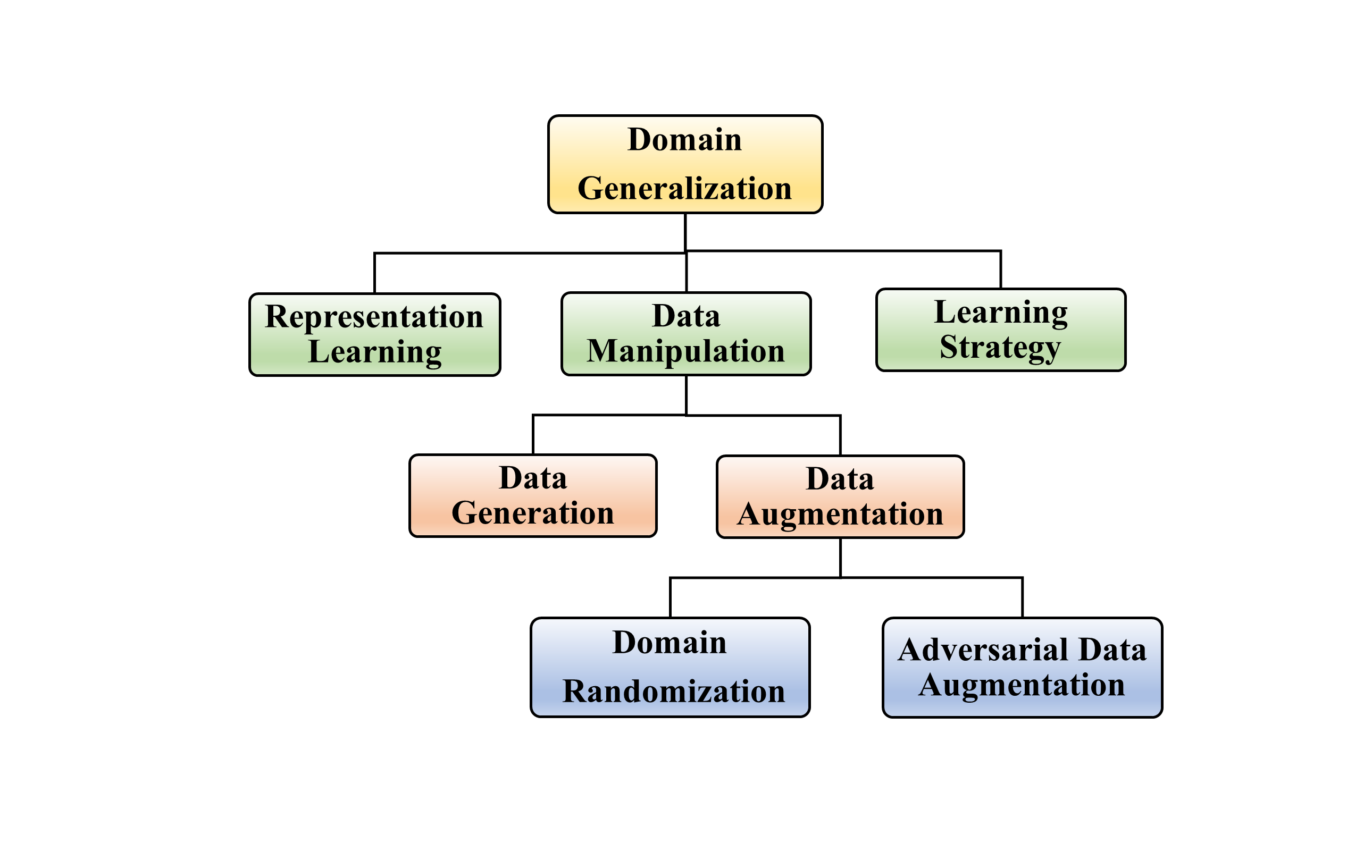}}
          %  \captionsetup{justification=centering}
		\caption{Relationship between Domain Randomization and Domain Generalization}
		\centering
		\label{fig:OD}
	\end{figure}

The introduction of domain generalization techniques has presented opportunities to enhance the effectiveness and precision of object identification models when applied to unfamiliar data. This study presents a mechanism that utilizes domain randomization to enhance the generalization capabilities of a model, without requiring any additional data collecting or annotation efforts. Furthermore, the empirical evaluation of the influence of transfer learning and mix-up augmentation is conducted for out-of-distribution data.

\section{Related Work}
Object detection comprises two sub-tasks: object localization and object categorization. Object detection has gained extensive utilization across various real-world applications, however,  given that different real-world detection tasks have distinct objectives and constraints, their difficulties can vary significantly. Apart from the common challenges encountered in other computer vision tasks, such as: handling object rotation, locating small objects, detecting occluded objects etc. Object detection encounters the additional challenge of detecting out-of-distribution or non-i.i.d data in real-world applications as the training procedure for most object detectors is based on an estimation of likelihood under the assumption of independent and identically distributed (i.i.d.) data \cite{10028728}. The resulting disparity in distribution between training and testing data, known as domain shift, negatively impacts model performance during deployment \cite{chen2021robust, hasan2021generalizable}. 
Domain adaptation (DA) \cite{chen2018domain, saito2019strong, xu2020exploring, li2020deep} includes methods such as feature regularization and adversarial training \cite{chen2018domain, hou2021informative, zhu2019adapting, saito2019strong, xu2020exploring} to tackle the challenge of domain shift by adjusting the parameters of a trained model using unlabeled data from the target domain(s) at the image, category, or object levels to achieve domain-invariant feature representation. However, in practice, target domain data is sparse and generally inaccessible. Unlike domain adaptation (DA), the objective of domain generalization (DG) is to acquire a consistent set of parameters that can effectively perform on previously unseen domains. Since it does not rely on data from the target domain or require a separate adaptation step \cite{li2017deeper}, it is a highly valuable setting against domain adaptation \cite{seemakurthy2022domain}.

 While theoretically, augmenting the quantity and diversity of training data can mitigate the impact of domain shift, it is important to note that annotating images remains a costly and time-consuming process. In addition to conventional augmentation techniques, domain randomization is a complementary approach to domain generalization as demonstrated in Figure \ref{fig:OD}, which involves generating new data that can simulate complex environments using the limited training samples available. \cite{wang2022generalizing}. 

Tobin et al.~\cite{tobin2017domain} proposed a Neural style transfer technique to randomize textures and positions of objects on the table in the simulation along with randomizing noise, light, and texture in the background to achieve better generalization for simulated datasets. Neural style transfer although effectively improves generalization for simulated datasets, yet it requires additional storage memory to store images and the effort required to select styles for image transformation \cite {shorten2019survey}. Mao et al. \cite{mao2021domain} merged virtual birds created from the 3D graphics toolset Blender with real-world backgrounds with sufficient variations in the environment to improve the model accuracy for bird detection. This work is different from \cite{mao2021domain} as we extract objects from a 2D real-world dataset instead of generating synthetic FOD objects from a 3D graphics toolset. A method to generate synthetic images for instance detection is proposed by \cite{dwibedi2017cut}. Dwibedi~et al. \cite{dwibedi2017cut} proposed to generate synthetic images by extracting objects from a baseline dataset i.e BigBird dataset \cite{singh2014bigbird} and blending on different backgrounds with patch-level realism. The baseline BigBird dataset \cite{singh2014bigbird} is a 3D large-scale dataset with 600 high-resolution images for each object. Each object is placed on a modest background and captured from 5 camera different angles. Corresponding depth images of baseline dataset are also available which are used as ground truth for training a ConvNet by \cite{dwibedi2017cut} for foreground/ background segmentation to generate segmentation mask. Extracted objects from baseline datasets are pasted on different backgrounds by applying blending techniques to include robustness from artifacts. The role of scene context in image augmentation is demonstrated in \cite{Dvornik2018OnTI,su2021context}.

The proposed method and problem at hand in this work differ greatly from \cite{dwibedi2017cut} in many aspects. FOD detection is an object detection problem and the FOD-A dataset is not a 3D dataset with each object captured either by a UAV or hand-held camera only. Images are comparatively of low resolution (300x300) with class imbalance. Due to light/day variations as well as weather variations in images captured from UAV and handheld cameras, the backgrounds of objects are not modest. The variation in viewpoint in FOD-A is also mostly by one angle, i.e., some images are captured at different heights from UAV. There are no depth images that can serve as ground truth for a ConvNet to predict foreground masks and subsequently segmentation masks. The background scenes are also in 300x300 resolution depicting practical runway challenges.

Based on the technique proposed by \cite{dwibedi2017cut}, we propose an algorithm for introducing randomized augmentation to generate synthetic images using FOD-A dataset which compliments the conclusion by Tobin et al for simulated datasets\cite{tobin2017domain} i.e., variation in training data style, forces the model to perceive real world as another variation, hence improving the generalization of state-of-the-art YOLO v5 object detector. Since the FOD-A dataset is representative of most of the real-world datasets with respect to resolution, class imbalance, and data paucity hence proposed methodology is applicable to most of the real-world datasets i.e., it is not restricted to large-scale 3D datasets for specific industrial problems. 

\section{Materials and Methods}
%In this work, we propose a two step approach to achieve a YOLOv5m based FOD detector, with comparable performance on source (in-domain data) and unseen (out-of-domain data):
In this study, we propose a two-step methodology for developing a YOLOv5m-based FOD detector. Our strategy aims to get similar performance levels on both source data (in-domain data) and unseen data (out-of-domain data):

% ------------------------------------------------------------------------------------------
\begin{itemize}
	\item Step 1: employ weak supervision to extract segmentation masks from the FOD-A dataset,
	\vspace{2mm}
	\item Step 2: generate synthetic images to enhance the dataset using randomized augmentations and backgrounds
\end{itemize}	
% ------------------------------------------------------------------------------------------

\subsection{Segmentation Masks Using BoxInst}

% ------------------------------------------------------------------------------------------
\begin{figure}[htb!]
	\centering
	\includegraphics[width=10.5 cm]{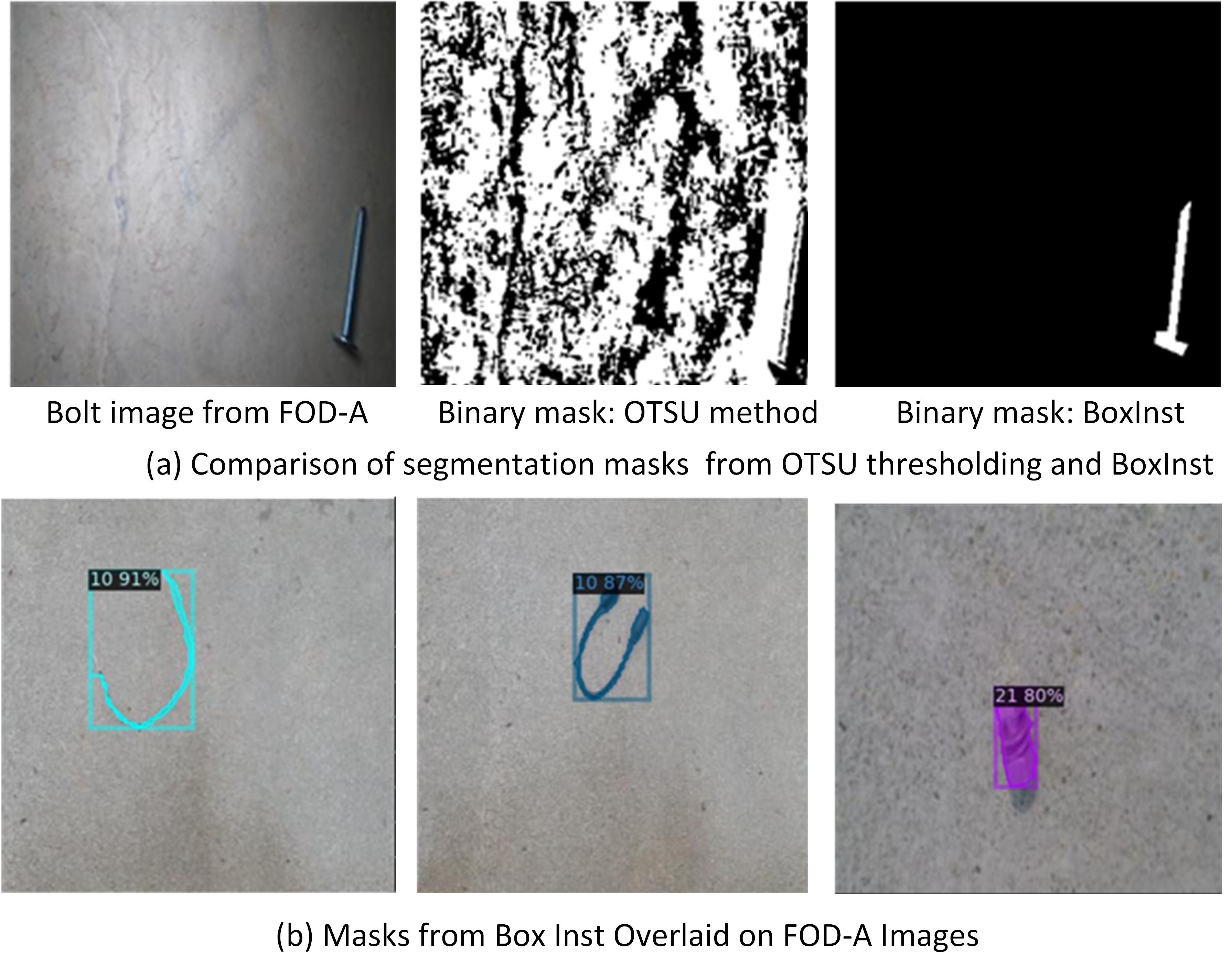}

	\caption{Comparison of segmentation mask from OTSU thresholding and BoxInst and selection of masks for SRIA}
	\label{fig:mask}
\end{figure}
% ------------------------------------------------------------------------------------------
Conventional approaches for segmenting objects from the background include trial and error, OTSU and HSV thresholding, however, these approaches upon empirical evaluation, failed to segment FOD in images with light variations as shown in Figure \ref{fig:mask}(a).The task of instance segmentation involves assigning labels to each pixel of an object. However, obtaining per-pixel annotations for segmentation masks is a time-intensive process \cite{tian2020conditional}\cite{bearman2016s}, as accurately outlining a single instance or object can take anywhere between 54 seconds \cite{jain2013predicting} and 79 seconds \cite{long2015fully}. 

Fully supervised instance segmentation requires ground truths for training in the form of {(M, c)}, where $M\in\ \{0, 1\}_{\text {H×W}}$  is the mask and $c \in\{1,2 \ldots. C\}$ is the object category. Since the FOD-A dataset has ground truths in the form of \{(B,c)\} where B is the bounding box surrounding the object of interest and per-pixel annotations are costly, we empirically evaluate weakly supervised learning to predict pixel-level masks as a first step. 

% ------------------------------------------------------------------------------------------
\begin{figure}[htb!]
	\centering
	\includegraphics[width=10.5 cm]{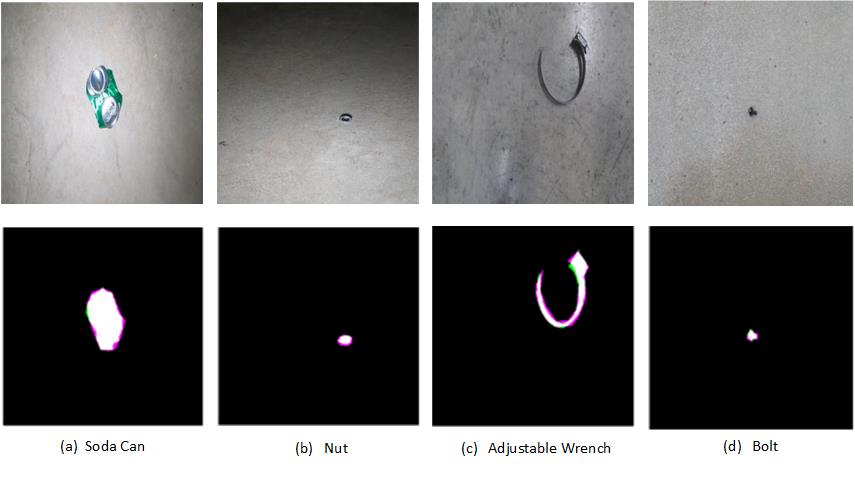}
	% where an .eps filename suffix will be assumed under latex, 
	% and a .pdf suffix will be assumed for pdflatex; or what has been declared
	% via \DeclareGraphicsExtensions.
 \captionsetup{justification=centering}
	\caption{Evaluating accuracy of BoxInst using Dice similarity coefficient.}
	\label{fig:dice}
\end{figure}
% ------------------------------------------------------------------------------------------

We initialize a BoxInst model\cite{tian2021boxinst} from the toolbox for Instance level recognition tasks \cite{tian2019adelaidet} with ImageNet pretrained weights and ResNet-50 backbone. For network parameters $\theta$, and predicted pixel level mask M ($\theta$), the network parameters are learned to minimize the loss function for the Box Inst model which is the sum of projection loss and pairwise affinity loss as given in Equation.

The projection loss term uses Dice loss to calculate the difference between the predicted mask M ($\theta$) and projection of B whereas pairwise loss is calculated on the assumption that proximal pixels with color similarity have the same label.     

\begin{equation}
L_{\text {mask }}=L_{\text {proj }}+L_{\text {pairwise }} 
\end{equation}
   
For pairwise similarity loss, we keep $\tau$ equal to $0.3$ and dilation at 2. Details of hyperparameters are given in Table \ref {table:boxinst}.

% ------------------------------------------------------------------------------------------
\begin{table}[t]
\caption{Hyperparameters for training BoxInst.}
\newcolumntype{C}{>{\centering\arraybackslash}X}
\begin{tabularx}{\textwidth}{CCCCCC}
\toprule
\textbf{Iterations	}	& \textbf{Batch Size}  & \textbf{Training Time}  &\textbf{Learning Rate}  &\textbf{Pairwise Loss}  &\textbf{Projection Loss}\\
\midrule
90000	& 8 & 45 hrs & 0.0001 & 0.007235 & 0.01657 \\
\bottomrule
\label{table:boxinst}
\end{tabularx}
\end{table}
% ------------------------------------------------------------------------------------------

FOD-A dataset is evaluated by the trained BoxInst model to get segmentation mask annotations. A sample of images from FOD-A are manually annotated using vgg Image Annotator \cite{dutta2016via} to compare the accuracy of BoxInst model with manual segmentation. Sørensen-Dice similarity coefficient as given in the following equation:
% ------------------------------------------------------------------------------------------
\ref{eqn:eq1}
\begin{equation}
\frac{2 *|X \cap Y|}{|X|+|Y|}
\label{eqn:eq1}
\end{equation}

% ------------------------------------------------------------------------------------------

which is calculated to validate the quality of segmentation masks. The value for Dice coefficients for images depicted in Figure \ref{fig:dice} are given in Table \ref{table:dice} which shows that the quality of generated masks is comparable to human annotators. The accuracy of Box-Inst is compared with the conventional OTSU technique and found to cater the light variations, unlike conventional approaches as shown in Figure \ref{fig:mask}(a). To select masks from each of the 31 classes of FOD-A, masks predicted by BoxInst are overlaid on images using a visualization tool from detectron2 \cite{wu2019detectron2} as shown in Figure \ref{fig:mask}(b)

\begin{table}[t] 
\caption{Sørensen-Dice similarity coefficient values for generated masks} 
\newcolumntype{C}{>{\centering\arraybackslash}X}
\begin{tabularx}{\textwidth}{CCCC}
\toprule
\textbf{Soda Can} & \textbf{Nut}  & \textbf {Adjustable Wrench}  &\textbf{Bolt} \\
\midrule
0.9375	& 0.7778  & 0.8269 & 0.8111 \\
\bottomrule
\label{table:dice}
\end{tabularx}
\end{table}

\subsection{Synthetic Images to Enhance Dataset}

In order to achieve a high level of realism at the patch level in synthetic images, data augmentation techniques are employed to produce diverse variances within the synthetic dataset. Upon doing a comprehensive examination of the FOD-A dataset, it becomes evident that while the collection encompasses a diverse range of potential foreign object debris (FOD) items found on runways, there is a limited degree of variance in the backdrops against which these things are depicted. In the actual operational environment, runway surfaces commonly exhibit cracks and other markings, such as white lines and yellow lines for taxiways. However, it should be noted that these specific features have not been accounted for in the FOD-A dataset. 

\subsubsection{Open Domain Background Images}
To generalize well and minimize false alarms, learning foreground/ background segregation and ignoring distractions is important for object detectors. For the FOD detection problem, we collected 29 images on a real runway which is the intended environment of the automated FOD detection model. To include the  practical challenges, we collect images depicting:

% -----------------------------------------------------------------------------------------
\begin{itemize}
	\item Runway yellow lines
	\vspace{1mm}
	\item Taxi way white lines
        \vspace{1mm}
        \item Runway Cracks
        \vspace{1mm}
        \item Tyre marks
        \vspace{1mm}
\end{itemize}
% -----------------------------------------------------------------------------------------

Figure~\ref{fig:combined} (a) depicts a sample of images collected to randomize the background for synthetic images.

\subsubsection{Synthetic Images Generation}

Let 'M' be the set of predicted masks from BoxInst model in Step 1 for category ‘C’ and 'B' be the set of backgrounds collected on a real runway. We synthesize new images for category ‘C’ in six batches by cutting the pixels from a randomly selected mask from ‘M’ and pasting them on a random background from B. For each batch, we select a unique combination of rotation “R”, scale  “S”, number of instances “I”, and Occlusion “O” augmentations with random values from a defined upper and lower limit.  Random truncation is applied in all iterations with a threshold of 0.25 ensuring that even after truncation, at least 25 percent object remains in the image. The synthesized images for category 'C' are the sum of images synthesized from all batches. For each category ’C’, a random number of synthetic images are generated in six batches with a maximum threshold of 65 images per batch. We do not paste objects on semantics-informed positions which is contrary to the proposal by \cite{georgakis2017synthesizing} and use no blending technique to paste objects on the background which is contrary to \cite{dwibedi2017cut} since many objects are small and the image resolution is not high, hence, blending in case of 2D low-resolution dataset might further compromise object feature information in the synthetic image

\vspace{2mm}
\begin{algorithm}[t]
\caption{Algorithm to generate synthetic images}
\begin{algorithmic}
\Require{Masks per class: M $\geq 1$, Backgrounds: B $\geq 1$}
\Ensure Rotation: R $\gets$ $-45^{\circ} $ to $45^{\circ} $ 
\State Scale: S $\gets$  $0.25$ to $0.6$;
\State Occlusion: O $\gets$ $0.6$;
\State Truncation: T $\gets$ $0.25$;
\State Instances: I $\gets$ $1$ to $6$;
%Attempts to synthesize $\gets$  $20$; 
 \State Class: C$\gets$ $0$ to $30$; 
\While{C $\leq 30$} \Comment{Start from Class 0}
  \State $B_1$=M*B*O*T; 
  \State $B_2$=M*B*R*S*O*T;
  \State $B_3$=M*B*I*T;
  \State $B_4$=M*B*S*R*I*T;
  \State $B_5$=M*B*R*T*O;
  \State $B_6$=M*N*S*T*O;
  \State Total images generated per class= sum ($B_1$ to $B_6$); 
  \State C$\gets$ C $+$ 1;
\EndWhile
\label{alg:wordy}
\end{algorithmic}
\end{algorithm}

\subsubsection{Synthetic Images Dataset}

\begin{figure}[t]
\centering
\includegraphics[width=10.5 cm]{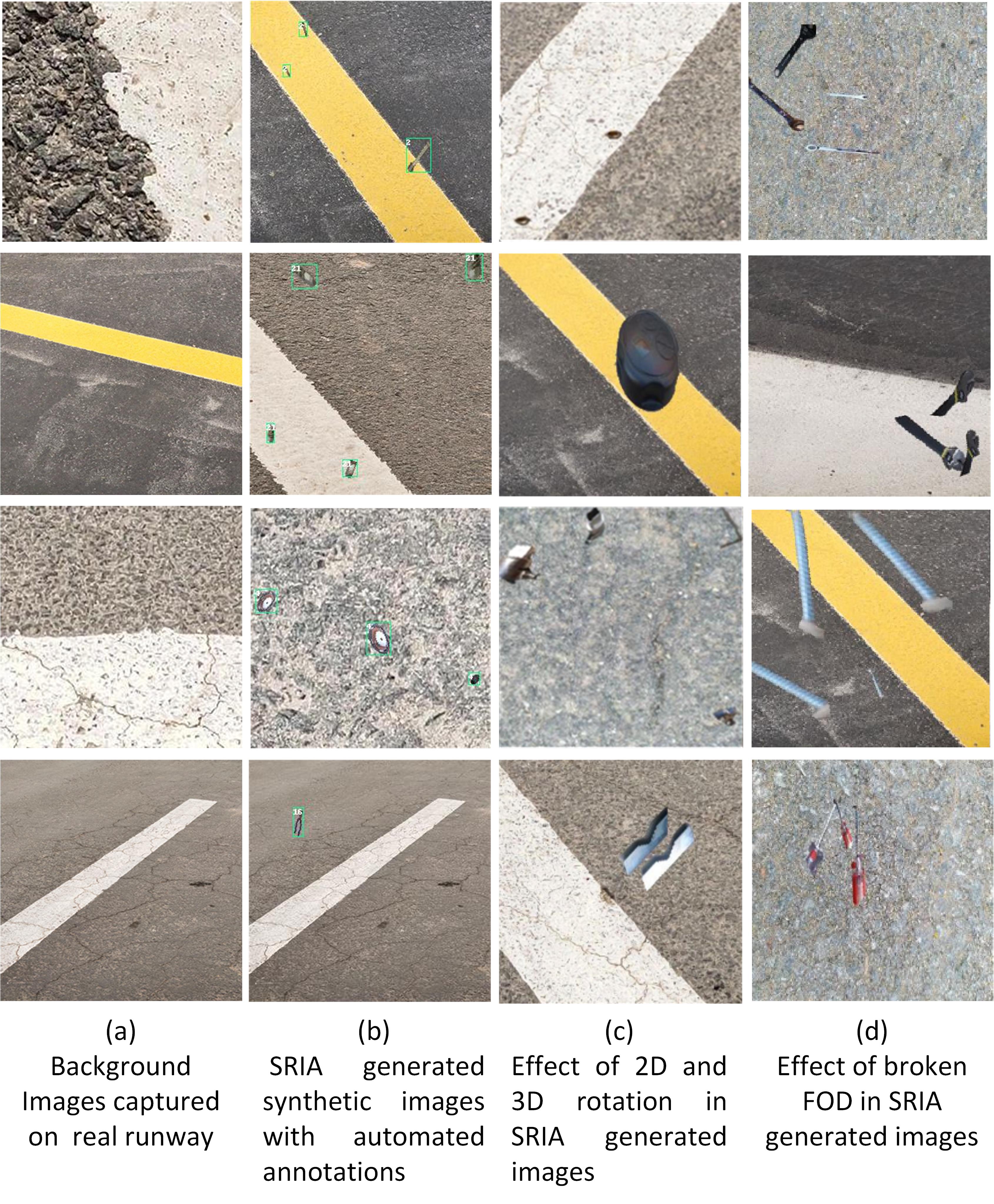}
\captionsetup{justification=centering}
\caption{SRIA generated synthetic images with automated annotations.
\label{fig:combined}}
\end{figure}

A total of $2227$ synthetic images are generated from $461$ extracted masks and $29$
background images. The statistics of generated images for each category are depicted in Table \ref{table:synthetic}. The generated images are evaluated for patch-level realism i.e., the placement of the extracted object looks realistic on the background without any context awareness. No blending techniques are applied. Figure \ref{fig:combined} (b) depicts a few samples from a generated
synthetic dataset with automatic annotations.
Viewpoint variation introduced by 2D rotation and 3D rotation in the synthetic
dataset looks realistic from human perception and is assumed to increase the feature learning capability. Some of the masks produced from BoxInst approach are partially formed. For the case of FOD detection, since FOD items can be broken and shape variants, the degraded / partially formed masks are also useful to increase the generalization of the model. Figure \ref{fig:combined} (c) depicts effect of 2D and 3D rotation whereas Figure \ref{fig:combined} (d) depicts images depicting broken FOD generated from a few partially formed masks obtained from the BoxInst approach.

% -------------------------------------------------------------------------
\begin{table}[t]
\small
\caption{Statistics of Synthetic Images.}
% \centering
\newcolumntype{C}{>{\centering\arraybackslash}X}
\begin{tabularx}{\textwidth}{CCC}
\toprule
\textbf{Class} & \textbf{No of masks extracted} & \textbf{No of images produced} \\
\toprule
Battery \           & 12 & 45    \\ 
Bolt washer        & 17 & 54  \\
Bolt                 & 16 & 75   \\ 
Clamp part           & 10 & 27    \\
Fuel cap     &  33 & 124    \\ 
Metal part & 14 & 58   \\ 
Nut   &  28 & 102  \\ 
Plastic part   &  27 & 105  \\ 
Rock    &  14 & 398   \\ 
Washer   &  12 & 48   \\ 
Wire   &  25 & 101   \\ 
Wrench   &  19 & 79   \\ 
Cutter   &  14 & 61   \\ 
Label   &  12 & 48   \\ 
Luggage tag   &  14 & 52   \\ 
Nail   &  25 & 94   \\ 
Pliers   &  24 & 98   \\ 
Metal sheet   &  1 & 38   \\ 
Hose   &  10 & 41   \\ 
Adjustable clamp   &  12 & 48   \\ 
Adjustable wrench   &  5 & 31   \\ 
Bolt nut   &  31 & 114   \\ 
Hammer   &  8 & 36   \\ 
Luggage part   &  8 & 33   \\ 
Paint chip   &  3 & 57   \\ 
Pen   &  11 & 45   \\ 
Screw   &  13 & 47   \\ 
Screw driver   &  10 & 40   \\ 
Soda can   &  14 & 56   \\ 
Wood   &  08 & 27   \\ 
Tape   &  11 & 45   \\ 
\textbf {Total}   &  \textbf{461} & \textbf{2227} \\ \bottomrule 
\label{table:synthetic} 
\end{tabularx}
\end{table}
% -----------------------------------------------------------------------------------------

% -----------------------------------------------------------------------------------------
% -----------------------------------------------------------------------------------------

\subsection{Out-of-Distribution (OOD) Dataset Preparation}
From the review of FOD datasets, it is established that there is no publicly available dataset for FOD detection except FOD-A which is used for training our model. 
Images from the OOD dataset are meant for FOD classification since there are no bounding box annotations. 10 common classes are identified and images are annotated to get ground truth for OOD data to evaluate the proposed model. 800 images from 10 common categories are annotated with bounding box annotations and termed as an OOD test set. The detail of categories with the number of images is tabulated in \ref{OOD} 

% -----------------------------------------------------------------------------------
\begin{table}[t]
\caption{Class wise detail of annotated OOD testset.}
\centering
\newcolumntype{C}{>{\centering\arraybackslash}X}
\begin{tabularx}{\textwidth}{CC}
\toprule
\textbf{Class} & \textbf{No of images} \\
\toprule
Clamp Part \           & 70    \\ 
Metal Part        & 93   \\ 
Pen                 & 48   \\ 
Nut           & 17    \\
Adjustable Clamp     &  44    \\
Wrench & 195   \\ 
Nail   &  52  \\ 
Rock   &  196  \\ 
Bolt    &  60   \\ 
Plastic Part   &  25   \\ 
\textbf {Total}   &  \textbf{800}    \\ 
\bottomrule
\label{OOD}
\end{tabularx} 
\end{table}
% -----------------------------------------------------------------------------------
      
\begin{figure}[t]
	\centerline{\includegraphics[width = 0.80 \textwidth]{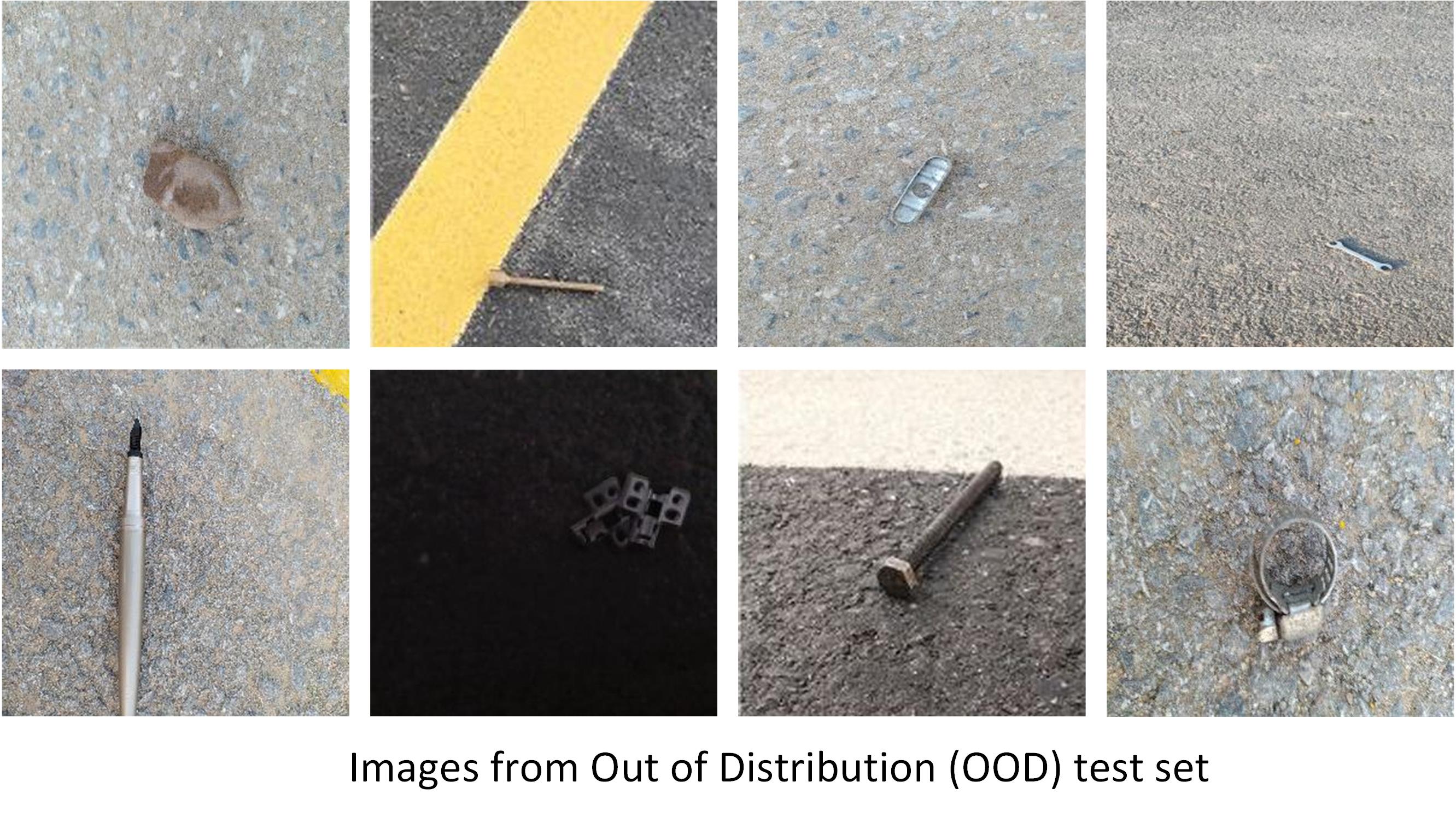}}
 \captionsetup{justification=centering}
	\caption{A sample of images from OOD dataset annotated for evaluation of proposed model}
	\centering
     \label{fig:sampleood}
\end{figure}

A sample of images annotated to evaluate performance of proposed model is demonstrated in Figure \ref{fig:sampleood}      
\noindent

\bigskip

\section{Experimentation}
 %For experimental setup, the  CPU is Intel I7 7700 K, the GPU is NVIDIA GeForce GTX 1070 and the Operating System is 64-bit Windows 10 x64-based processor.
 In the experimental setup, the central processing unit (CPU) utilized is an Intel I7 7700 K, the graphics processing unit (GPU) employed is the NVIDIA GeForce GTX 1070, and the operating system deployed is a 64-bit Windows 10 system running on an x64-based processor. 
%We set the following hyperparameters: For the ResNet50 training, we choose: a
%learning rate of 10��3 , cross-entropy loss function, SGD optimizer, exponential learning
%rate decay with gamma set to 0.95, and weight %decay 2  10��3.
YOLOv5m is trained using transfer learning by starting training from pre-trained weights on the COCO dataset for 70 epochs. The train-test split is kept the same as the baseline FOD-A model. The model is trained for four cases:

\begin{itemize}
	
	\item YOLOv5m using source+synthetic data with default YOLOv5 augmentations(SRIA)
	\item YOLOv5m using synthetic data with default YOLOv5 augmentations (SRIA$_S$)
	\item YOLOv5m using source+synthetic data with added augmentations and mix-up method (SRIA w/augmentation)
	\item YOLOv5m using source data only with added augmentation and mix-up method (YOLOv5m w/augmentation) 
	
\end{itemize}

YOLOv5m w/augmentation and SRIA w/augmentation are trained by modifying augmentations during training. Image rotation and image shear are introduced during training in a range of -$45$ to +$45$ deg. The image perspective fraction is set to 0.001. The main added augmentation for improving generalization is mix-up augmentation for which we have selected a probability of 0.2.

\section{Results and Discussion}
\label{results}
Evaluation parameters for object detection algorithms generally include mean average precision (mAP), and precision–recall (PR) curve. 
Precision and Recall are calculated on the basis of true positive (TP), true negative (TN), false negative (FN), and false positive (FP) from a comparison of model predicted class and true class of the objects.

\begin{equation}
	P=\frac{\mathrm{TP}}{(\mathrm{TP}+\mathrm{FP})} ; \quad \quad 
	P=\frac{\mathrm{TP}}{( all\:detections)}
\end{equation}

\begin{equation}
	 R=\frac{\mathrm{TP}}{(\mathrm{TP}+\mathrm{FN})} ; \quad \quad
	 R=\frac{\mathrm{TP}}{( all\: ground\:truths)} 
\end{equation}

From Eq. 1 and Eq. 2, It is clear that P represents the percentage of correct positive predictions whereas R refers to the percentage of true positive detected from all ground truths. The precision–recall curve (PR curve)  determines the detection accuracy of the model. A large area under the PR curve indicates high AP for the particular class \cite{roy2022fast}. Mean Average Precision (mAP) refers to the mean AP over all IoU thresholds or all classes.

We quantitatively assess the efficacy of our proposed methodology, SRIA, by evaluating its performance with regard to the mean Average Precision (mAP). Additionally, we undertake a qualitative evaluation to corroborate the effectiveness of SRIA. Subsequently, we employ the post-hoc explainability technique known as Eigen Class Activation Mapping (Eigen CAM) to gain insights into the qualitative outcomes of our approach. This comprehensive evaluation process allows us to both measure the quantitative performance and interpret the qualitative aspects of SRIA's performance.

\subsection{Quantitative Results}

Annotated OOD dataset is evaluated for:
\begin{itemize}
	\item YOLOv5m baseline model (Trained on source images only)
	\item SRIA$_S$ (Transfer Learning, trained on synthetic images only)
	\item SRIA (Transfer Learning, trained on source + synthetic images)
	\item SRIA w/augmentation (Transfer Learning, rotate, shear and mix-up augmentation)
	\item YOLOv5 w/augmentation(Transfer Learning,rotate, shear and mix-up augmentation) 
\end{itemize}

% ------------------------------------------------------------------------------
\begin{table}[t]
\caption{Quantitative comparison of SRIA for FOD-A Dataset.}
\centering
\newcolumntype{C}{>{\centering\arraybackslash}X}
\begin{tabularx}{\textwidth}{CCC}
\toprule
\textbf{Method} & \textbf{mAP (0.5-0.95)} & \textbf{Environment Specific} \\
\toprule
SSD \cite{munyer2021fod}           & 79.6  & No  \\ 
YOLOv3 \cite{munyer2021fod}        & 69.7  & No  \\ 
CenterNet            & 85.9 & No  \\ 
Scaled YOLOv4 P6         & 83.1  & No  \\ 
Scaled YOLOv4 P5        & 82.5  & No  \\ 
Outer ViT \cite{munyer2022foreign} & 82.7  & Yes \\ 
YOLOv4-csp w/Augmentation \cite{noroozi2023towards}  & 90.12 & Yes \\ 
YOLOv5m baseline    & 91.1 & No \\ 
\textbf{Ours(SRIA)}                                         & \textbf {91.9}  & \textbf {Yes} \\ \bottomrule
\label{sourceresult}
\end{tabularx} 
\end{table}
% -----------------------------------------------------------------------------------------

% -------------------------------------------------------------------------------------------
\begin{table}[t]
\caption{Quantitative comparison of SRIA for small test set.}
\centering
\newcolumntype{C}{>{\centering\arraybackslash}X}
\begin{tabularx}{\textwidth}{CCCCC}
\toprule
\textbf{Method} & \textbf{mAP (0.5)} &  \textbf{mAP (0.5-0.95)} & \textbf{Precision} & \textbf{Recall} \\ 
\toprule
YOLOv5m baseline          & 0.993  & 0.873  & 0.974  & 0.986  \\ 
YOLOv5m w/augmentation          & 0.994  & 0.882  & 0.985  & 0.98 \\ 
SRIA$_S$       & 0.779  & 0.555 &  0.751   & 0.733 \\
SRIA w/augmentation                & 0.994 & 0.893 & 0.97    &  0.987 \\
\textbf{Ours(SRIA)}                 & 0.994 & \textbf{0.895} & 0.985    & 0.989 \\ \bottomrule
\label{smallresult}
\end{tabularx}
\end{table}
% -------------------------------------------------------------------------------------------
% -------------------------------------------------------------------------------------------
\begin{table}[t]
\caption{Quantitative comparison of SRIA for OOD test set.}
\centering
\newcolumntype{C}{>{\centering\arraybackslash}X}
\begin{tabularx}{\textwidth}{CCCCC}
\toprule
\textbf{Method} & \textbf{mAP (0.5)} &  \textbf{mAP (0.5-0.95)} & \textbf{Precision} & \textbf{Recall} \\ \midrule
YOLOv5m baseline          & 0.418  & 0.255  & 0.602  & 0.362  \\
YOLOv5m w/augmentation          & 0.62  & 0.41  & 0.567  & 0.567 \\
SRIA$_S$        & 0.72  & 0.433 &  0.718   & 0.681 \\
SRIA w/augmentation                & 0.742 & 0.484 & 0.712    &  0.71 \\
\textbf{Ours(SRIA)}                 & \textbf{0.926} & \textbf{0.6} & \textbf{0.903}   & \textbf{0.831} \\\bottomrule
\label{oodresult}
\end{tabularx} 
\end{table}
% -------------------------------------------------------------------------------------------

Quantitative evaluation results are depicted in Table \ref{sourceresult} for the source (FOD-A) dataset, Table \ref{smallresult} for small FOD test set (manually selected small FOD images from FOD-A dataset), and Table \ref {oodresult} for OOD test set.Quantitative evaluation results from the YOLOv5m baseline model, our proposed approach SRIA, YOLOv5m w/ augmentation, SRIA w/augmentation shows that our proposed approach SRIA (trained with source + synthetic images), outperforms other evaluated models in precision, recall, and mAP by a large margin and specifically improves detection accuracy for small FODs from 87\% to 89.5\% as depicted in Table \ref {smallresult} and OOD test set from 41\% to 92 \% as depicted in Table \ref {oodresult}. SRIA also outperforms all compared methodologies in the existing literature including SSD \cite {munyer2021fod}, YOLOv3 \cite {munyer2021fod},  Outer Vit \cite{munyer2022foreign}, YOLOv4-csp w/Augmentation \cite{noroozi2023towards} in terms of mAP as depicted in Table \ref{sourceresult}

The contribution of mix up augmentation in improving generalization of YOLOv5m is also evident from our experimental results depicted in  Table \ref{oodresult}. We find that the detection accuracy of YOLOv5 baseline model improves from 41 \%  to  62 \% for out-of-distribution (OOD) testset for the YOLOv5m w/augmentation model which is trained by employing mix up augmentation. 

We also find out that scene context does not affect the FOD detection problem when natural backgrounds are used  which differs from the findings of \cite{dvornik2019importance} \cite {su2021context}. We also find that using natural backgrounds with possible practical problems lead to significant improvement in FOD detection results for real-time FOD detection problem. This finding complements the study by \cite{nesteruk2023cisa}

\subsection{Qualitative Results} 
\begin{figure}[hbt!]
	\centerline{\includegraphics[width = 0.80 \textwidth]{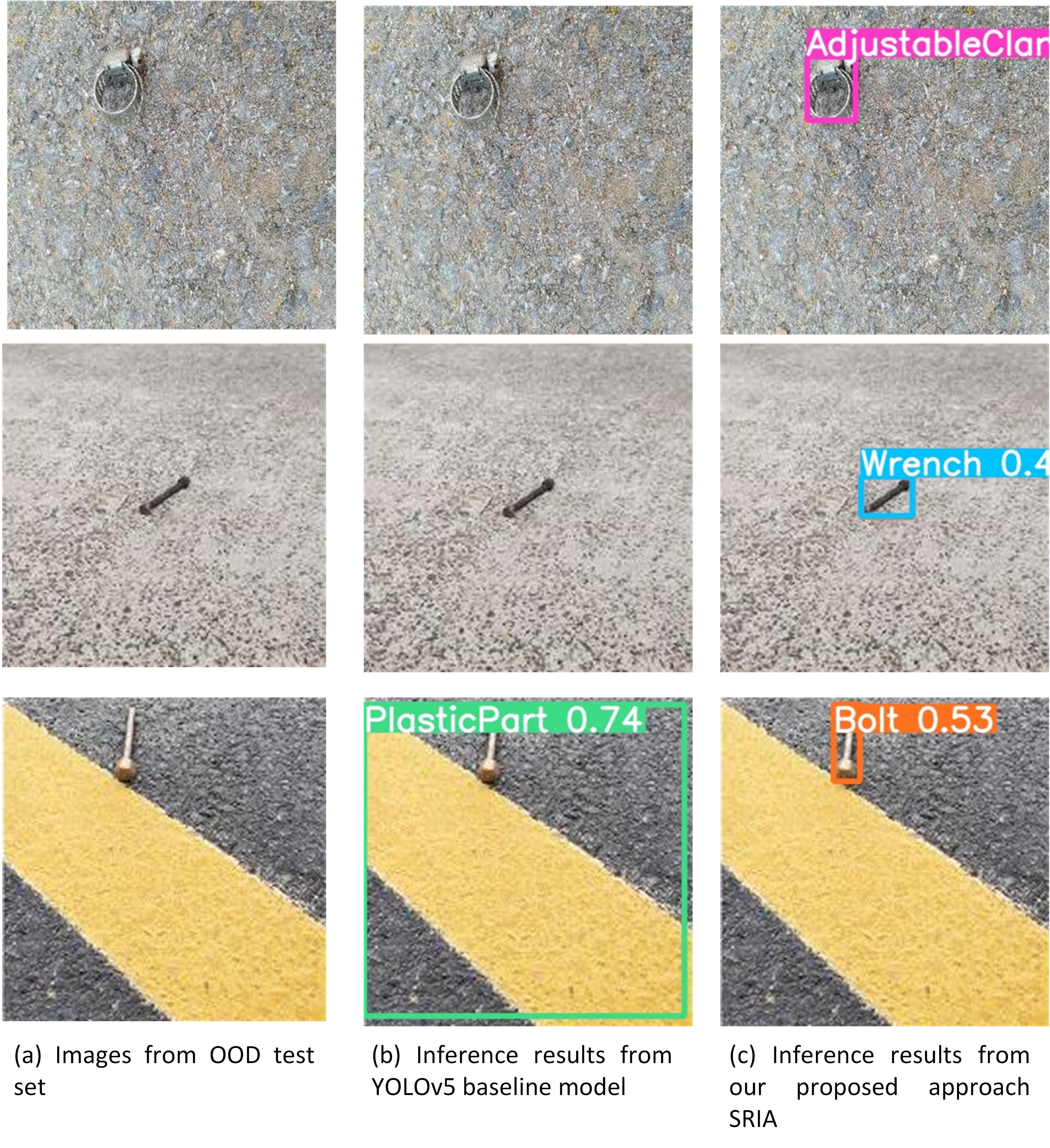}}
 \captionsetup{justification=centering}
	\caption{Comparing generalization results for YOLOv5m Baseline model and our proposed approach SRIA for images from OOD dataset}
	\centering
	\label{fig:result}
\end{figure}

Predictions from the proposed model and baseline model are plotted on the images for qualitative evaluation of the generalization of our proposed technique. We have evaluated images from three sources including images from the OOD test set, images from field trials of FOD systems \cite{metalonroadboth} \cite{wrenchplastic}, and images from Mechanical Tools Classification Dataset\cite{Mechanical Tool Classification Dataset} with complicated backgrounds having common categories with FOD-A dataset. The qualitative results for the OOD test set are demonstrated in Figure \ref{fig:result}. 

\begin{figure}[hbt!]
	\centerline{\includegraphics[width = 0.80 \textwidth]{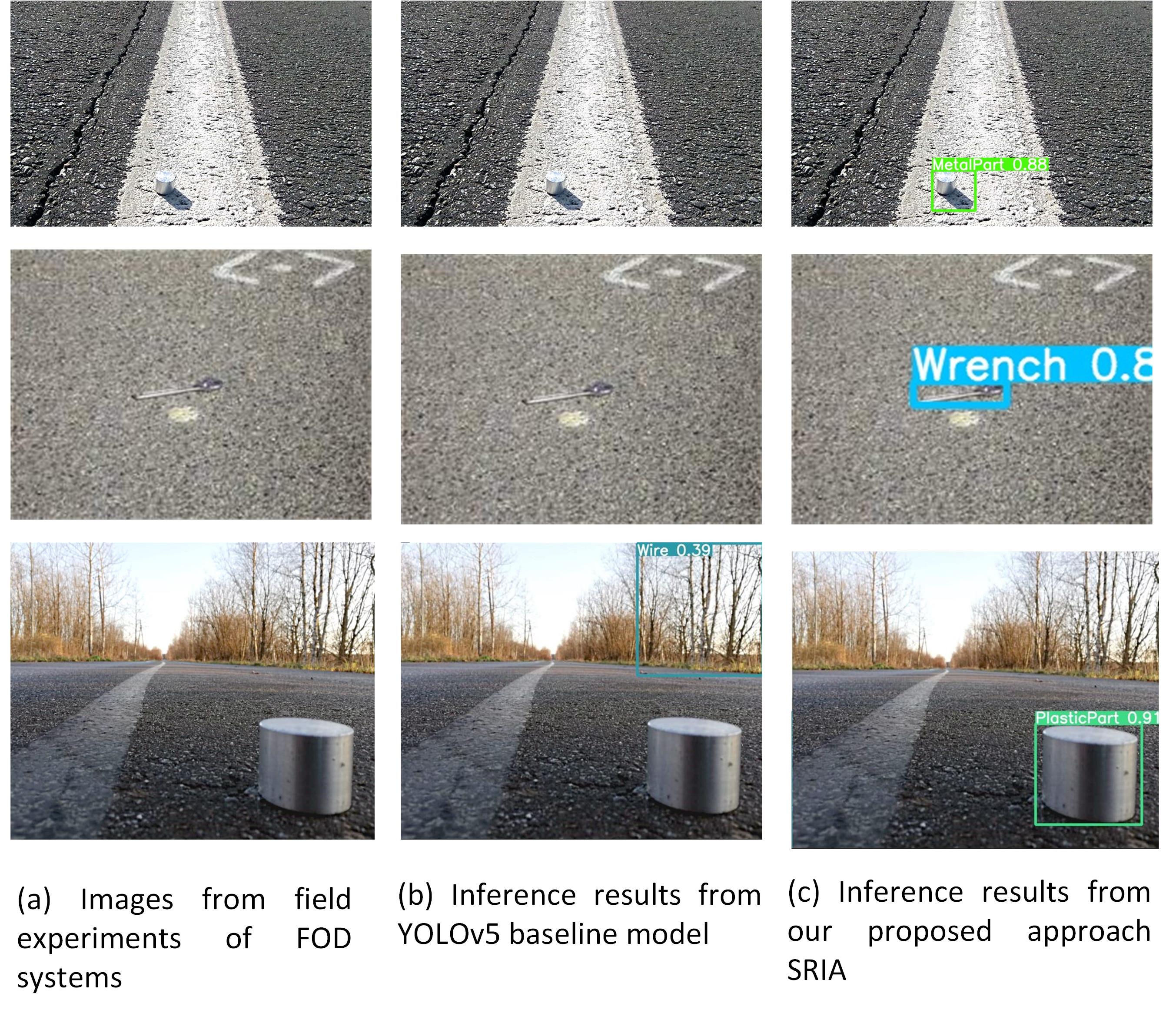}}
 \captionsetup{justification=centering}
	\caption{Comparing generalization results for YOLOv5m Baseline model and our proposed approach SRIA for images  from field experiments of FOD systems}
	\centering
	\label{fig:internet}
\end{figure}

SRIA model is able to correctly detect and classify most of the objects in the OOD test set whereas the YOLOv5m baseline model has much more false positives and missed detections. Figure \ref{fig:internet} shows the inference results from SRIA compared with the inference results from the YOLOv5 baseline model for images from actual field trials of FOD systems. Our proposed model SRIA is able to accurately detect and classify the out-of-distribution images (never seen by the model before), where the YOLOv5m baseline results in missed detections and false positives.

% -------------------------------------------------------------------------------------------
% -------------------------------------------------------------------------------------------

\begin{figure}[hbt!]
	\centerline{\includegraphics[width = 0.80 \textwidth]{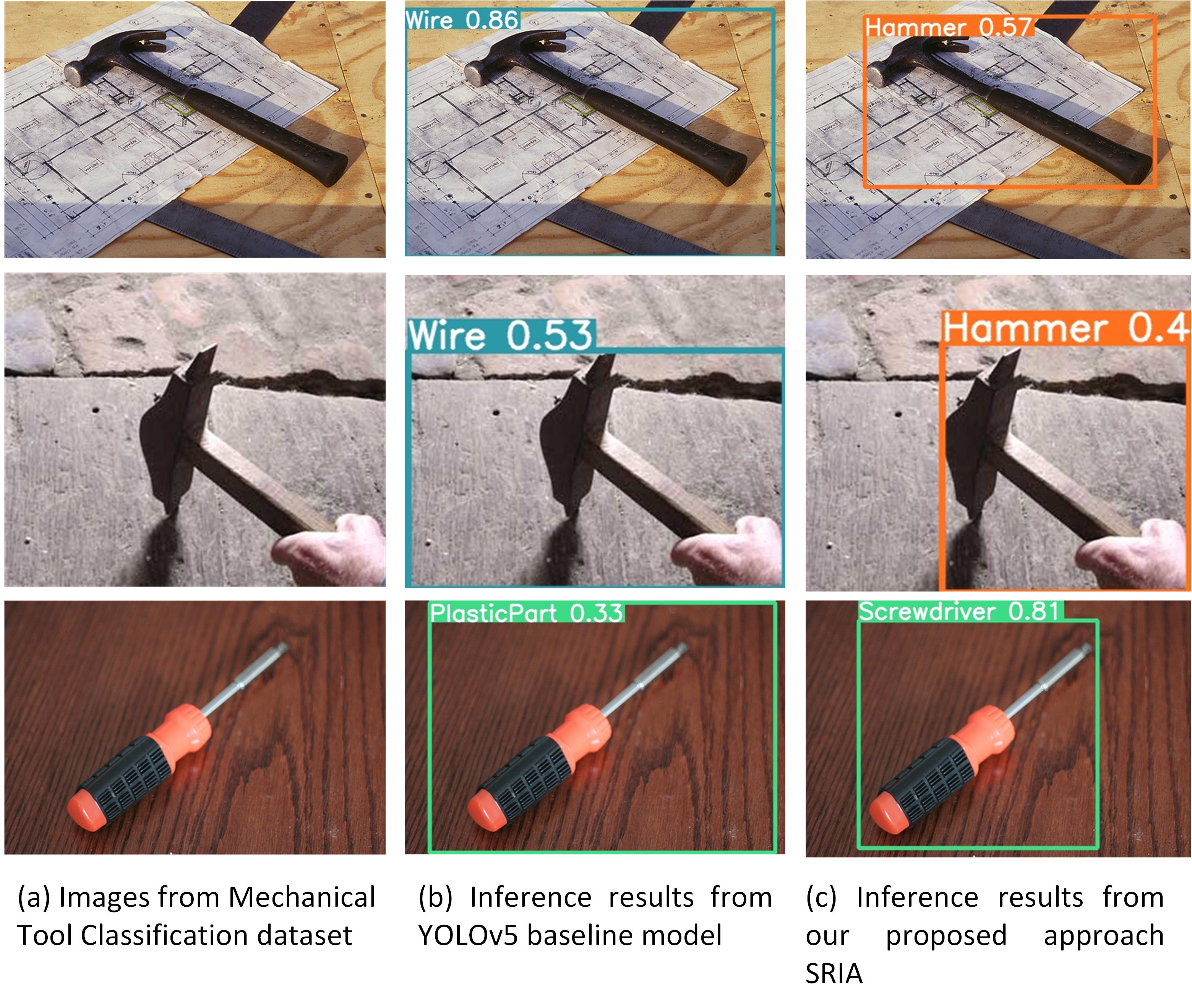}}
 \captionsetup{justification=centering}
	\caption{Comparing generalization results for YOLOv5m Baseline model and our proposed approach SRIA for images from Mechanical Tool Classification Dataset}
	\centering
	\label{fig:mechanical}
\end{figure}

For further qualitative evaluation, Images from the Mechanical Tools Classification Dataset\cite{Mechanical Tool Classification Dataset} are inferred from SRIA and YOLOv5 baseline model. Results are shown in Figure \ref{fig:mechanical}. From the qualitative results, it is evident that the proposed SRIA approach focuses the model on learning features of objects and eliminates the focus on background which has resulted in better generalization for out-of-distribution images as evident from Figure \ref{fig:internet}, and  Figure \ref{fig:mechanical}.

% -------------------------------------------------------------------------------------------
% -------------------------------------------------------------------------------------------

Explainability refers to the causal factors behind decision-making. Post-hoc explainability techniques aim to explain why the model has inferred a result, while transparency design techniques integrate constraints of explainability during model design. Out of many DNN approaches proposed for FOD detection, only a few \cite{noroozi2023towards}have focused on the explainability of the model to learn the causal factors behind the model's detection/ classification of FOD. 

% -------------------------------------------------------------------------------------------
\begin{figure}[hbt!]
	\centerline{\includegraphics[width = 0.80 \textwidth]{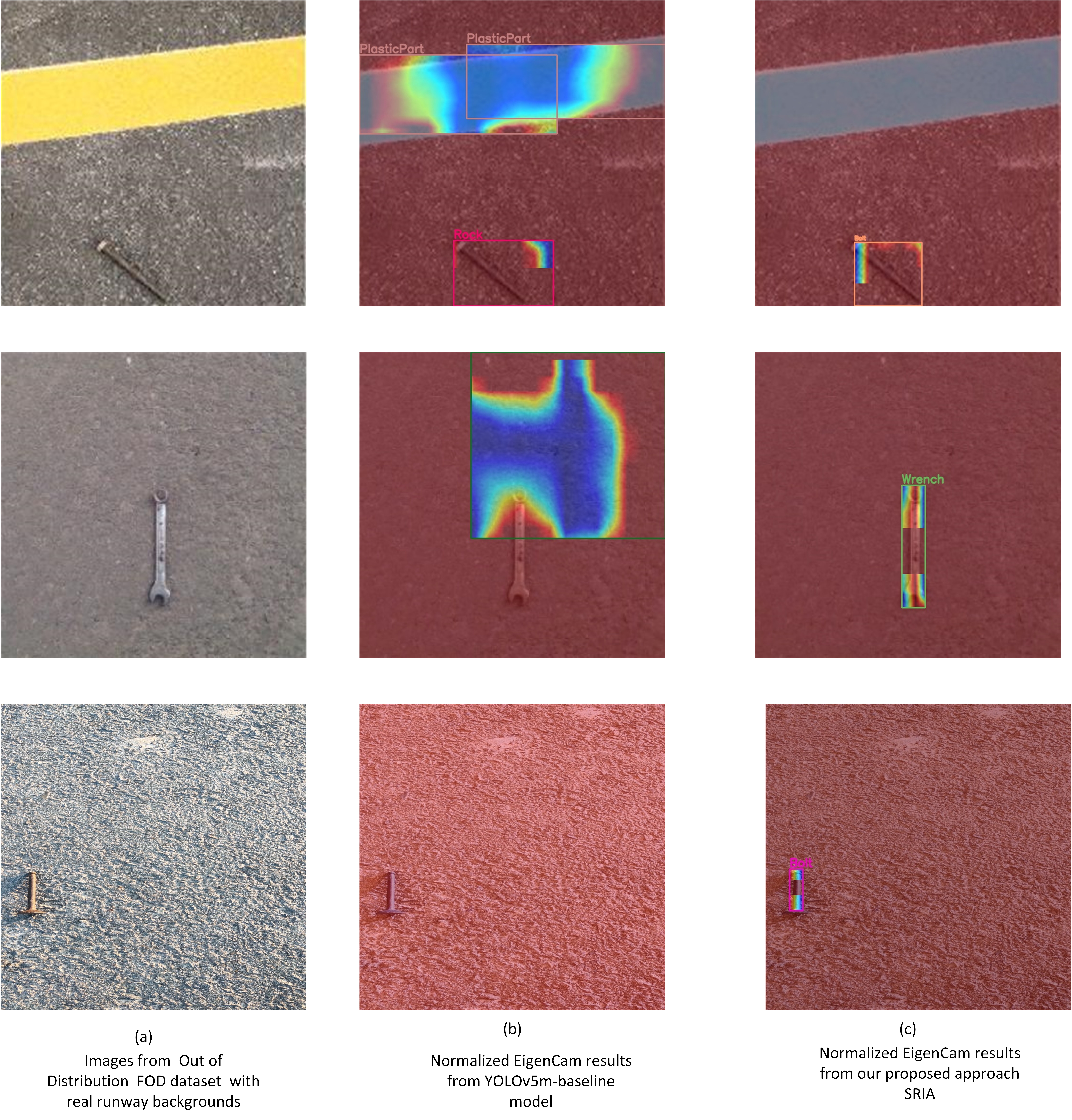}}
 \captionsetup{justification=centering}
	\caption{EigenCam results for YOLOv5m Baseline model and our proposed approach SRIA to compare the generalization of models for OOD images}
	\centering
	\label{fig:explain}
\end{figure}
% -------------------------------------------------------------------------------------------

To understand the explainability of the model, we compare inference from the YOLOv5 baseline and our proposed approach SRIA using posthoc explainability techniques Eigen Cam \cite{muhammad2020eigen} for images from OOD test set. Eigen Class Activation Mapping (Eigen-CAM) is a post-hoc method that aims to interpret the knowledge acquired by a model from data or explain the reasons behind its poor performance in a given task by calculating and displaying the primary components of the acquired features/representations from the convolutional layers through visualization. Eigen Cam implementation \cite{jacobgilpytorchcam} with target layer: -2 is used to analyze the interpretations from the model. We normalize the CAM in the range [0,1] inside every bounding box. Normalization takes the results outside the bounding box as zero.  Figure \ref{fig:explain} (a)  depicts three images Out-of-Distribution (OOD) dataset with real runway backgrounds. The inference results from EigenCam for YOLOv5m and our proposed SRIA approach are depicted in Figure \ref{fig:explain} (b) and (c) respectively. The Eigen Cam results help in understanding the reason for the degraded performance of the YOLOv5m baseline model and the strength of our proposed approach SRIA. First, as no actual runway backgrounds are available in the FOD-A dataset when YOLOv5m baseline model infers an OOD image captured on an actual runway, it predicts runway yellow lines as plastic parts and misclassifies 'bolt' as 'rock' as shown in Figure \ref{fig:explain} (b)-Top. Second, due to less variation in backgrounds and relatively less variety in image capturing angles and lights in the FOD-A dataset, the YOLOv5 baseline model gives false negatives for FODs in OOD images under varying lights and orientations as evident from Figure \ref{fig:explain} (b)-Middle as it focuses more on background instead of wrench features. Similarly YOLOv5m baseline fails to detect bolts on rough runway surfaces with light variation as depicted in Figure \ref{fig:explain} (b)-Bottom. The strength of our proposed method is also revealed through Eigen Cam results SRIA is robust to varying runway backgrounds and correctly classifies 'bolt', and 'wrench' based on the specific features as shown in Figure \ref{fig:explain} (c). 

Our proposed approach SRIA introduces random variations during training through synthetic images, hence it focuses the model more towards the FOD features and does not get distracted from variations in background, lights, and object orientations. The improvement in generalization is achieved through domain randomization with no significant increment in dataset size.

%%%%%%%%%%%%%%%%%%%%%%%%%%%%%%%%%%%%%%%%%%
\section{Conclusion}

In this research endeavor, we have introduced a data-driven methodology for the detection of foreign object debris (FOD) within the context of the FOD-A dataset. This approach offers cost-effectiveness, universality in its applicability across diverse airport environments, and adaptability for the identification of hitherto unrecognized FOD instances. Our study underscores the considerable potential of synthetic imagery in augmenting the generalization capabilities of the YOLOv5 model. By means of our two-step technique, we generate synthetic images that, upon assimilation into the training dataset, encourage the model to prioritize salient features while dismissing inconsequential variations. Furthermore, our investigation underscores the merits of mix-up augmentation in enhancing the model's generalization prowess. Additionally, we make a significant contribution to future research endeavors by providing comprehensive annotations for a real-world runway test dataset, encompassing ten prevalent FOD categories in conjunction with the FOD-A dataset. Our innovative approach not only outperforms extant computer vision-based methodologies documented in the existing literature but also excels in the identification of both familiar and previously unknown foreign objects.

Due to the myriad potential manifestations of foreign object debris (FOD) on airport runways, it is neither economically viable nor practically feasible to construct an exhaustive dataset encompassing every conceivable FOD instance. Such an undertaking would incur substantial costs and still fall short in accommodating the emergence of novel FOD types. Consequently, the future trajectory of research should be directed towards several key avenues. Firstly, the augmentation of backgrounds can be pursued by integrating imagery sourced from aircraft simulators. Furthermore, the incorporation of real-world distracting objects, such as taxiway lights and runway end indicator lights (REIL), within synthetic images can enhance the realism of the dataset. Additionally, amalgamating objects from diverse sources can be explored as a strategy to elevate classification accuracy. To circumvent the necessity of retraining the entire model each time a novel FOD class or type is introduced to the dataset, it is prudent to investigate task incremental learning methods, such as "learning without forgetting" (LwF) \cite{li2017learning}. This avenue holds promise for future research endeavors. The augmentation algorithm presented in this paper harbors the potential for broader applicability to other data domains where predefined background scenarios prevail. Subsequent research inspired by this study could delve into assessing the extent to which this method can be extrapolated to analogous scenarios, thereby exploring its generalizability.

\backmatter

\section*{Declarations}

\begin{itemize}
\item Funding {This research received no external funding.}
\item Conflict of interest {The authors declare no conflict of interest.} 
\item Availability of data and materials {The FOD-A dataset analysed during the current study is available in the [FOD-UNOmaha] repository, [https://github.com/FOD-UNOmaha/FOD-data. The data presented in this study are available on request from the corresponding author. The data are not publicly available due to privacy reasons}
\item Authors' contributions { ``Conceptualization, J.F ; methodology, J.F; software, J.F; validation, N.A and M.K.K; investigation, J.F.; resources, A.S ; data curation, J.F and M.I.S ; writing original draft preparation, J.F.; writing---review and editing, N.A; visualization, M.K.K; supervision, N.A; project administration, A.S , M.I.S. All authors have read and agreed to the published version of the manuscript.''.}
\end{itemize}

\noindent

\
%%===========================================================================================%%


\begin{thebibliography}{999}
% Reference 1
\bibitem{zhang2020semantics}Zhang, P., Lan, C., Zeng, W., Xing, J., Xue, J. \& Zheng, N. Semantics-guided neural networks for efficient skeleton-based human action recognition. {\em Proceedings Of The IEEE/CVF Conference On Computer Vision And Pattern Recognition}. pp. 1112-1121 (2020)

\bibitem{liu2021survey}Liu, Y., Sun, P., Wergeles, N. \& Shang, Y. A survey and performance evaluation of deep learning methods for small object detection. {\em Expert Systems With Applications}. \textbf{172} pp. 114602 (2021)

\bibitem{shen2021towards}Shen, Z., Liu, J., He, Y., Zhang, X., Xu, R., Yu, H. \& Cui, P. Towards out-of-distribution generalization: A survey. {\em ArXiv Preprint ArXiv:2108.13624}. (2021)

\bibitem{zaidi2022survey}Zaidi, S., Ansari, M., Aslam, A., Kanwal, N., Asghar, M. \& Lee, B. A survey of modern deep learning based object detection models. {\em Digital Signal Processing}. \textbf{126} pp. 103514 (2022)

\bibitem{zou2023object}Zou, Z., Chen, K., Shi, Z., Guo, Y. \& Ye, J. Object detection in 20 years: A survey. {\em Proceedings Of The IEEE}. (2023)


\bibitem{chauhan2020review} Chauhan, T., Goyal, C., Kumari, D. \& Thakur, A. A review on foreign object debris/damage (FOD) and its effects on aviation industry. {\em Materials Today: Proceedings}. \textbf{33} pp. 4336-4339 (2020)

\bibitem{chen2021robust}Chen, X., Xie, C., Tan, M., Zhang, L., Hsieh, C. \& Gong, B. Robust and accurate object detection via adversarial learning. {\em Proceedings Of The IEEE/CVF Conference On Computer Vision And Pattern Recognition}. pp. 16622-16631 (2021)

\bibitem{hasan2021generalizable}Hasan, I., Liao, S., Li, J., Akram, S. \& Shao, L. Generalizable pedestrian detection: The elephant in the room. {\em Proceedings Of The IEEE/CVF Conference On Computer Vision And Pattern Recognition}. pp. 11328-11337 (2021)

\bibitem{zhong2021fod}Zhong, J., Gou, X., Shu, Q., Liu, X. \& Zeng, Q. A fod detection approach on millimeter-wave radar sensors based on optimal vmd and svdd. {\em Sensors}. \textbf{21}, 997 (2021)

\bibitem{yonemoto2018two}Yonemoto, N., Kohmura, A., Futatsumori, S., Morioka, K. \& Kanada, N. Two dimensional radar imaging algorithm of bistatic millimeter wave radar for FOD detection on runways. {\em 2018 International Topical Meeting On Microwave Photonics (MWP)}. (2018)

\bibitem{cao2016foreign}
X.~Cao, G.~Gong, M.~Liu, and J.~Qi, ``Foreign object debris detection on
airfield pavement using region based convolution neural network,'' in
\emph{2016 International Conference on Digital Image Computing: Techniques
and Applications (DICTA)}.\hskip 1em plus 0.5em minus 0.4em\relax IEEE, 2016,
pp. 1--6.

\bibitem{cao2017detecting}
X.~Cao, Y.~Gu, and X.~Bai, ``Detecting of foreign object debris on airfield
pavement using convolution neural network,'' in \emph{LIDAR Imaging Detection
and Target Recognition 2017}, vol. 10605.\hskip 1em plus 0.5em minus
0.4em\relax International Society for Optics and Photonics, 2017, p. 1060536.

\bibitem{cao2018region}
X.~Cao, P.~Wang, C.~Meng, X.~Bai, G.~Gong, M.~Liu, and J.~Qi, ``Region based
cnn for foreign object debris detection on airfield

\bibitem{hong2018experiment}Hong, J., Kang, M., Kim, Y., Kim, M. \& Hong, G. Experiment on automatic detection of airport debris (FOD) using EO/IR cameras and radar. {\em Journal Of Advanced Navigation Technology}. \textbf{22}, 522-529 (2018)

\bibitem{yuan2020research}
Z.-D. Yuan, J.-Q. Li, Z.-N. Qiu, and Y.~Zhang, ``Research on fod detection
system of airport runway based on artificial intelligence,'' in \emph{Journal
of Physics: Conference Series}, vol. 1635, no.~1.\hskip 1em plus 0.5em minus
0.4em\relax IOP Publishing, 2020, p. 012065.

\bibitem{lin2014microsoft}
T.-Y. Lin, M.~Maire, S.~Belongie, J.~Hays, P.~Perona, D.~Ramanan,
P.~Doll{\'a}r, and C.~L. Zitnick, ``Microsoft : Common objects in context,''
in \emph{European conference on computer vision}.\hskip 1em plus 0.5em minus
0.4em\relax Springer, 2014, pp. 740--755.

\bibitem{liu2018fod}
Y.~Liu, Y.~Li, J.~Liu, X.~Peng, Y.~Zhou, and Y.~L. Murphey, ``Fod detection
using densenet with focal loss of object samples for airport runway,'' in
\emph{2018 IEEE Symposium Series on Computational Intelligence (SSCI)}.\hskip
1em plus 0.5em minus 0.4em\relax IEEE, 2018, pp. 547--554

\bibitem{munyer2021fod}
T.~Munyer, P.-C. Huang, C.~Huang,\& X.~Zhong, ``Fod-a: A dataset for foreign object debris in airports,'' \emph{arXiv preprint arXiv:2110.03072 }, 2021.

\bibitem{papadopoulos2021uav}
E.~Papadopoulos and F.~Gonzalez, ``UAV and AI application for runway foreign
object debris (fod) detection,'' in \emph{2021 IEEE Aerospace Conference
(50100)}.\hskip 1em plus 0.5em minus 0.4em\relax IEEE, 2021, pp. 1--8.

\bibitem{GC} Glenn-Jocher. Available online: https://github.com/ultralytics/yolov5 (accessed on 27 Feb, 2022).

\bibitem{deng2009imagenet}Deng, J., Dong, W., Socher, R., Li, L., Li, K. \& Fei-Fei, L. Imagenet: A large-scale hierarchical image database. {\em 2009 IEEE Conference On Computer Vision And Pattern Recognition}. pp. 248-255 (2009)

\bibitem{russakovsky2015imagenet}
O.~Russakovsky, J.~Deng, H.~Su, J.~Krause, S.~Satheesh, S.~Ma, Z.~Huang,
A.~Karpathy, A.~Khosla, M.~Bernstein \emph{et~al.}, ``Imagenet large scale
visual recognition challenge,'' \emph{International journal of computer
vision}, vol. 115, no.~3, pp. 211--252, 2015.

\bibitem{seemakurthy2022domain}Seemakurthy, K., Fox, C., Aptoula, E. \& Bosilj, P. Domain Generalisation for Object Detection. {\em ArXiv Preprint ArXiv:2203.05294}. (2022)

\bibitem{torralba2011unbiased}Torralba, A. \& Efros, A. Unbiased look at dataset bias. {\em CVPR 2011}. pp. 1521-1528 (2011)

\bibitem{yue2019domain}Yue, X., Zhang, Y., Zhao, S., Sangiovanni-Vincentelli, A., Keutzer, K. \& Gong, B. Domain randomization and pyramid consistency: Simulation-to-real generalization without accessing target domain data. {\em Proceedings Of The IEEE/CVF International Conference On Computer Vision}. pp. 2100-2110 (2019)

\bibitem{wang2022generalizing}Wang, J., Lan, C., Liu, C., Ouyang, Y., Qin, T., Lu, W., Chen, Y., Zeng, W. \& Yu, P. Generalizing to unseen domains: A survey on domain generalization. {\em IEEE Transactions On Knowledge And Data Engineering}. (2022)

\bibitem{knoll2019assessment}Knoll, F., Hammernik, K., Kobler, E., Pock, T., Recht, M. \& Sodickson, D. Assessment of the generalization of learned image reconstruction and the potential for transfer learning. {\em Magnetic Resonance In Medicine}. \textbf{81}, 116-128 (2019)

\bibitem{tian2021boxinst}Tian, Z., Shen, C., Wang, X. \& Chen, H. Boxinst: High-performance instance segmentation with box annotations. {\em Proceedings Of The IEEE/CVF Conference On Computer Vision And Pattern Recognition}. pp. 5443-5452 (2021)

\bibitem{wu2019detectron2}Wu, Y., Kirillov, A., Massa, F., Lo, W. \& Girshick, R. Detectron2. (https://github.com/facebookresearch/detectron2,2019)

\bibitem{tian2019adelaidet}Tian, Z., Chen, H., Wang, X., Liu, Y. \& Shen, C. AdelaiDet: A Toolbox for Instance-level Recognition Tasks. (https://git.io/adelaidet,2019)

\bibitem{tobin2017domain}Tobin, J., Fong, R., Ray, A., Schneider, J., Zaremba, W. \& Abbeel, P. Domain randomization for transferring deep neural networks from simulation to the real world. {\em 2017 IEEE/RSJ International Conference On Intelligent Robots And Systems (IROS)}. pp. 23-30 (2017)

\bibitem{shorten2019survey}Shorten, C. \& Khoshgoftaar, T. A survey on image data augmentation for deep learning. {\em Journal Of Big Data}. \textbf{6}, 1-48 (2019)

\bibitem{mao2021domain}Mao, X., Chow, J., Tan, P., Liu, K., Wu, J., Su, Z., Cheong, Y., Ooi, G., Pang, C. \& Wang, Y. Domain randomization-enhanced deep learning models for bird detection. {\em Scientific Reports}. \textbf{11}, 639 (2021)


\bibitem{dwibedi2017cut}Dwibedi, D., Misra, I. \& Hebert, M. Cut, paste and learn: Surprisingly easy synthesis for instance detection. {\em Proceedings Of The IEEE International Conference On Computer Vision}. pp. 1301-1310 (2017)

\bibitem{singh2014bigbird}Singh, A., Sha, J., Narayan, K., Achim, T. \& Abbeel, P. Bigbird: A large-scale 3d database of object instances. {\em 2014 IEEE International Conference On Robotics And Automation (ICRA)}. pp. 509-516 (2014)

\bibitem{Dvornik2018OnTI}Dvornik, N., Mairal, J. \& Schmid, C. On the Importance of Visual Context for Data Augmentation in Scene Understanding. {\em IEEE Transactions On Pattern Analysis And Machine Intelligence}. \textbf{43} pp. 2014-2028 (2018)

\bibitem{su2021context}Su, Y., Sun, R., Lin, G. \& Wu, Q. Context Decoupling Augmentation for Weakly Supervised Semantic Segmentation.  (2021)

\bibitem{tian2020conditional}Tian, Z., Shen, C. \& Chen, H. Conditional convolutions for instance segmentation. {\em European Conference On Computer Vision}. pp. 282-298 (2020)

\bibitem{bearman2016s}Bearman, A., Russakovsky, O., Ferrari, V. \& Fei-Fei, L. What’s the point: Semantic segmentation with point supervision. {\em European Conference On Computer Vision}. pp. 549-565 (2016)

\bibitem{long2015fully}Long, J., Shelhamer, E. \& Darrell, T. Fully convolutional networks for semantic segmentation. {\em Proceedings Of The IEEE Conference On Computer Vision And Pattern Recognition}. pp. 3431-3440 (2015)

\bibitem{dutta2016via}Dutta, A., Gupta, A. \& Zissermann, A. wu2019detectron2 Image Annotator (VIA). (http://www.robots.ox.ac.uk/ wu2019detectron2/software/via/,2016)

\bibitem{jain2013predicting}Jain, S. \& Grauman, K. Predicting sufficient annotation strength for interactive foreground segmentation. {\em Proceedings Of The IEEE International Conference On Computer Vision}. pp. 1313-1320 (2013)

\bibitem{dvornik2019importance}Dvornik, N., Mairal, J. \& Schmid, C. On the importance of visual context for data augmentation in scene understanding. {\em IEEE Transactions On Pattern Analysis And Machine Intelligence}. \textbf{43}, 2014-2028 (2019)

\bibitem{nesteruk2023cisa}Nesteruk, S., Zherebtsov, I., Illarionova, S., Shadrin, D., Somov, A., Bezzateev, S., Yelina, T., Denisenko, V. \& Oseledets, I. CISA: Context Substitution for Image Semantics Augmentation. {\em Mathematics}. \textbf{11}, 1818 (2023)

\bibitem{roy2022fast}
A.~M. Roy, R.~Bose, and J.~Bhaduri, ``A fast accurate fine-grain object
  detection model based on yolov4 deep neural network,'' \emph{Neural Computing
  and Applications}, pp. 1--27, 2022.
 
\bibitem{jacobgilpytorchcam}Gildenblat, J. \& Contributors PyTorch library for CAM methods. (GitHub,2021), https://github.com/jacobgil/pytorch-grad-cam

\bibitem{wrenchplastic} FOD detection research. Available online: https://www.internationalairportreview.com/article/1192/foreign-object-debris-fod-detection-research/
 
\bibitem{metalonroadboth} Success of Field Trial Experiment of 76 GHz High Precision Foreign Object Debris Detection Radar for Runway in UUTO. %Available online: https://elva-1.com/news_events/a40169

\bibitem{muhammad2020eigen}Muhammad, M. \& Yeasin, M. Eigen-cam: Class activation map using principal components. {\em 2020 International Joint Conference On Neural Networks (IJCNN)}. pp. 1-7 (2020)

\bibitem{munyer2022foreign}Munyer, T., Brinkman, D., Zhong, X., Huang, C. \& Konstantzos, I. Foreign Object Debris Detection for Airport Pavement Images based on Self-supervised Localization and Vision Transformer. {\em ArXiv Preprint ArXiv:2210.16901}. (2022)

\bibitem{noroozi2023towards}Noroozi, M. \& Shah, A. Towards optimal foreign object debris detection in an airport environment. {\em Expert Systems With Applications}. \textbf{213} pp. 118829 (2023)

\bibitem{10028728}Zou, Z., Chen, K., Shi, Z., Guo, Y. \& Ye, J. Object Detection in 20 Years: A Survey. {\em Proceedings Of The IEEE}. \textbf{111}, 257-276 (2023)

\bibitem{li2017learning}Li, Z. \& Hoiem, D. Learning without forgetting. {\em IEEE Transactions On Pattern Analysis \& Machine Intelligence}. \textbf{40}, 2935-2947 (2017)

\bibitem {toolsisolated} Available online at https://www.shutterstock.com/search/tools-isolated

\bibitem{Mechanical Tool Classification Dataset} Salman ibne unus. November 2021. Mechanical Tools Classification Dataset, Version:8.24. Retrieved September 2022 from https://www.kaggle.com/datasets/salmaneunus/mechanical-tools-dataset?resource=download.

\bibitem{chen2018domain}Chen, Y., Li, W., Sakaridis, C., Dai, D. \& Van Gool, L. Domain adaptive faster r-cnn for object detection in the wild. {\em Proceedings Of The IEEE Conference On Computer Vision And Pattern Recognition}. pp. 3339-3348 (2018)

\bibitem{georgakis2017synthesizing}Georgakis, G., Mousavian, A., Berg, A. \& Kosecka, J. Synthesizing training data for object detection in indoor scenes. {\em ArXiv Preprint ArXiv:1702.07836}. (2017)

\bibitem{hou2021informative}Hou, L., Zhang, Y., Fu, K. \& Li, J. Informative and consistent correspondence mining for cross-domain weakly supervised object detection. {\em Proceedings Of The IEEE/CVF Conference On Computer Vision And Pattern Recognition}. pp. 9929-9938 (2021)

\bibitem{zhu2019adapting}Zhu, X., Pang, J., Yang, C., Shi, J. \& Lin, D. Adapting object detectors via selective cross-domain alignment. {\em Proceedings Of The IEEE/CVF Conference On Computer Vision And Pattern Recognition}. pp. 687-696 (2019)

\bibitem{saito2019strong}Saito, K., Ushiku, Y., Harada, T. \& Saenko, K. Strong-weak distribution alignment for adaptive object detection. {\em Proceedings Of The IEEE/CVF Conference On Computer Vision And Pattern Recognition}. pp. 6956-6965 (2019)

\bibitem{xu2020exploring}Xu, C., Zhao, X., Jin, X. \& Wei, X. Exploring categorical regularization for domain adaptive object detection. {\em Proceedings Of The IEEE/CVF Conference On Computer Vision And Pattern Recognition}. pp. 11724-11733 (2020)

\bibitem{li2020deep}Li, W., Li, F., Luo, Y., Wang, P. \& Others Deep domain adaptive object detection: A survey. {\em 2020 IEEE Symposium Series On Computational Intelligence (SSCI)}. pp. 1808-1813 (2020)

\bibitem{li2017deeper}Li, D., Yang, Y., Song, Y. \& Hospedales, T. Deeper, broader and artier domain generalization. {\em Proceedings Of The IEEE International Conference On Computer Vision}. pp. 5542-5550 (2017)




\end{thebibliography}
\end{document}